\documentclass{article}
\usepackage[preprint]{colm2026_conference}

\usepackage{microtype}
\usepackage{hyperref}
\usepackage{url}
\usepackage{booktabs}
\usepackage{amsmath}
\usepackage{amssymb}
\usepackage{subcaption}
\usepackage{float}
\usepackage{graphicx}
\usepackage{caption}
\usepackage{etoolbox}
\usepackage{fancyhdr}
\usepackage{lineno}
\usepackage{enumitem}
\usepackage{xcolor}
\usepackage{colortbl}
\usepackage{multirow}
\usepackage[table]{xcolor}
\usepackage{lipsum}
\usepackage{makecell}
\usepackage{wrapfig}
\definecolor{lightgray}{RGB}{240,240,240}
\definecolor{lightyellow}{RGB}{255,248,220}
\definecolor{darkblue}{RGB}{180,210,255}

\definecolor{gainG}{HTML}{0B7A2A}
\definecolor{lossR}{HTML}{C0141B}
\definecolor{fadeG}{HTML}{C8C8C8}
\definecolor{linkblue}{rgb}{0, 0, 0.5}
\hypersetup{colorlinks=true, citecolor=linkblue, linkcolor=linkblue, urlcolor=linkblue}

\definecolor{redDark}{HTML}{C0392B}
\definecolor{redMid}{HTML}{E07060}
\definecolor{redLight}{HTML}{F2B4AE}
\definecolor{grnLight}{HTML}{B7DFB0}
\definecolor{grnMid}{HTML}{52BE80}
\definecolor{grnDark}{HTML}{1E7D34}
\definecolor{neutralGray}{HTML}{F8F8F8}

\usepackage{tcolorbox}

\title{\fullmethod: Recovering Linguistic Ability in Vision-Language Models via Selective Cross-Modal Distillation}

\author{Patrick Amadeus Irawan, Erland Hilman Fuadi, Shanu Kumar, \\
\textbf{Alham Fikri Aji, Yova Kementchedjhieva} \\ 
Mohamed bin Zayed University of Artificial Intelligence\\
\texttt{\{patrick.irawan, yova.kementchedjhieva\}@mbzuai.ac.ae}
}

\newcommand{\fullmethod}{\textsc{LinguDistill }}

\newcommand{\vanillavlm}{\texttt{nanoVLM}}
\newcommand{\basevlm}{\texttt{nanoVLM-full}}
\newcommand{\selectvlm}{\texttt{nanoVLM-lang}}
\newcommand{\distill}{\texttt{distill-full}}

\newcommand{\adaptdistill}{\textsc{LinguDistill}}

\makeatletter
\newcommand\mytopfigure{%
  \begin{center}%
    \includegraphics[width=0.95\linewidth]{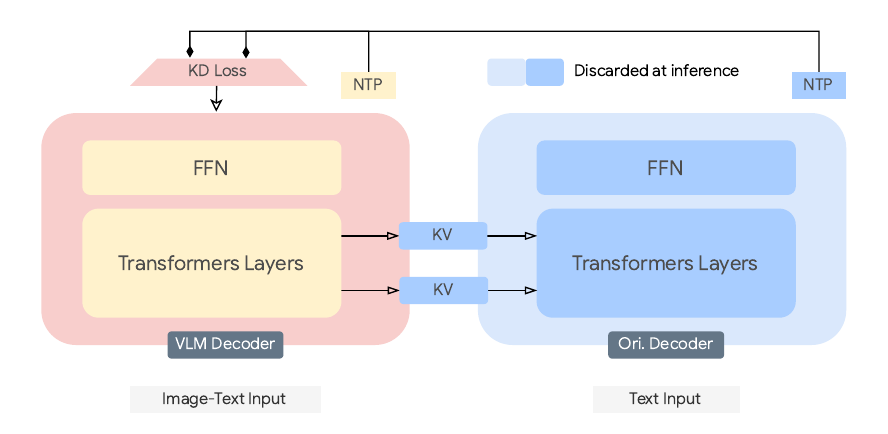}%
    \captionof{figure}{\fullmethod training setup. The VLM decoder (student, pink) and frozen pretrained LM (teacher, blue) communicates by reusing the student's representation via layer-wise KV sharing. The teacher's vision-aware outputs are distilled back into the student via a selective KD. After training, the teacher is removed, resulting in linguistically-improved VLM with zero additional modules.}
    \label{fig:hook}%
  \end{center}%
}
\apptocmd\@maketitle{{\mytopfigure\par}}{}{}
\makeatother

\begin{document}

\ifcolmsubmission
\linenumbers
\fi

\maketitle

\begin{abstract}
Turning a pretrained language model (LM) into a vision-language model (VLM) through multimodal fine-tuning often erodes its native language ability, a form of catastrophic forgetting that shows up even on text-only tasks. This loss is hard to undo with further fine-tuning, and existing remedies add adapters or alignment modules that increase architectural complexity and inference cost. We propose \fullmethod, an adapter-free knowledge distillation method that uses the original frozen LM as the teacher during multimodal post-training. To let a text-only teacher supervise vision-conditioned outputs, we introduce layer-wise KV-cache sharing, which exposes the teacher to the student's multimodal representations without changing either architecture. We then apply distillation selectively, on language-heavy data only, so the teacher restores linguistic ability while the student keeps its visual grounding on document and OCR tasks. \fullmethod recovers the language and knowledge performance lost during multimodal fine-tuning, matching the original VLM on average over text-only benchmarks (ARC, HellaSwag) and exceeding it on ScienceQA, while keeping vision-heavy performance close to standard fine-tuning. Since the teacher is dropped after training, the final model adds no parameters and no inference cost. More broadly, our results show that a model's own pre-adaptation backbone is a practical teacher for undoing forgetting, suggesting a simple recipe for keeping language ability intact as models are extended to new modalities.
\end{abstract}

\newpage

\section{Introduction}

Building multimodal models requires aligning different modalities into a shared representation space. In vision-language models (VLMs), this is typically achieved through fusion-based architectures \citep{liu2023llava,bai2025qwen3vltechnicalreport,li2025multimodalalignmentfusionsurvey,zhao2026unifiedmultimodalunderstandinggeneration}, which train the backbone language model (LM) to understand non-text modalities, or through earlier bridging paradigms that keep the LM frozen \citep{li2023blip2,alayrac2022flamingovisuallanguagemodel}. While the former generally demonstrates stronger performance, it comes at a cost. Adapting pretrained LM to multimodal settings leads to degradation in their native language capabilities, even pronounced when evaluated on purely textual inputs \citep{linguisticforgetting,wang2025crossmodalknowledgedistillationspeech}. This phenomenon, often attributed to modality interference or catastrophic forgetting, suggests that multimodal adaptation perturbs the linguistic priors encoded in the original backbone.


A natural approach is to further fine-tune the model on language-heavy tasks. However, prior work suggests that this is insufficient to fully recover language capability, as multimodal adaptation causes a representation shift and cross-modal interference that are not easily reversed through lightweight fine-tuning \citep{wang2025crossmodalknowledgedistillationspeech}. As a result, language ability remains degraded even after further adaptation. To address this, prior work has explored several alternative directions. Bridging approaches, such as BLIP-style alignment, keep the LM largely frozen and introduce intermediate alignment modules to preserve its linguistic priors \citep{wang2025iaa,wang2024cogvlm,zhang2024llamaadapter}. Alternatively, subspace or weight-space methods attempt to recover lost capabilities by manipulating model parameters, for example through weight interpolation, task arithmetic, or selective parameter restoration \citep{wortsman2022wiseft,ilharco2023editingmodelstaskarithmetic,zhu2024modeltailor,zhang2024wings}. However, these approaches either rely on additional alignment modules, require large-scale training from scratch, or assume that desired capabilities can be cleanly isolated and restored in parameter space, which may not hold across models and modalities.

In this paper, we propose \fullmethod, a cross-modal knowledge distillation method to restore language capability in VLMs. We treat the frozen pretrained LM as a teacher and the VLM as a student, which is optimized using a standard distillation objective. A key challenge is that the teacher only operates on text, so its outputs are not aligned with vision-conditioned generation. To address this, we introduce a KV-cache sharing mechanism between the teacher and student decoders. This allows the LM to access the same multimodal context as the VLM and produce vision-aware supervision signals. After training, the teacher is removed, resulting in a standard VLM with improved language capability and no additional parameters or inference cost.

We organize our study around three research questions:
\begin{itemize}[leftmargin=*]
    \item \textbf{RQ1 (Degradation).} Does continued multimodal fine-tuning erode the linguistic capability of a VLM, and can restricting training to language-heavy data alone prevent it?
    \item \textbf{RQ2 (Recovery).} Can the original frozen, text-only LM provide useful vision-aware supervision through KV-cache sharing, and thereby recover the lost linguistic capability?
    \item \textbf{RQ3 (Trade-off).} What does distillation from a text-only teacher cost on visually grounded tasks, and does routing the distillation signal by data type contain that cost?
\end{itemize}

Our contributions are as follows:
\begin{itemize}
    \item We propose an adapter-free selective distillation method that recovers native language capability in VLMs by using the frozen original LM backbone as a teacher.
    \item We introduce a layer-wise KV-cache sharing architecture that lets a frozen LM attend to multimodal context with no extra modules, and is discarded after training.
    \item We show that \fullmethod recovers, and on several benchmarks exceeds, the language and knowledge performance of the original VLM, while selective routing keeps the cost on document and OCR tasks far below that of uniform distillation.
\end{itemize}

\section{Related Work}
\label{sec:related}

\paragraph{Preserving Language Capability in Multimodal Models.}
Fine-tuning a pretrained language model on multimodal data degrades its language capability, a form of catastrophic forgetting documented in several recent studies \citep{zhai2023investigating,linguisticforgetting}.
The dominant approach to this problem freezes the language backbone entirely and inserts learnable modules to handle visual input.
Frozen \citep{tsimpoukelli2021frozen} trains only a vision encoder whose outputs are fed to a fully frozen GPT-J.
BLIP-2 \citep{li2023blip2} adds a Q-Former between image features and the frozen LM.
CogVLM \citep{wang2024cogvlm} injects visual experts as parallel attention branches at every transformer layer.
LLaMA-Adapter \citep{zhang2024llamaadapter} adds zero-initialised attention adapters to the top layers.
IAA \citep{wang2025iaa} inserts inner adapters that alternate visual and textual processing at varying depths.
These methods preserve language capability by never updating the LM, but they all add parameters that increase inference cost.
Our method instead recovers language capability after standard fine-tuning, using the original LM as a distillation teacher and adding no parameters to the final model.

\paragraph{Cross-Modal Knowledge Distillation.}
Knowledge distillation \citep{hinton2015distilling} trains a student to match the soft output distribution of a teacher via KL divergence, combined with a hard cross-entropy loss on ground-truth labels.
Cross-modal KD extends this to settings where teacher and student operate on different input modalities.
\citet{xue2023modalityfocusing} show that cross-modal KD works because the student learns modality-general features shared across modalities.
C2KD \citep{huo2024c2kd} provides a general framework for bridging the modality gap between teacher and student across audio-visual, image-text, and RGB-depth pairs.
In the LLM era, \citet{wang2025crossmodalknowledgedistillationspeech} distil a text-only LLM into a speech LLM, transferring linguistic capability across modalities.
In vision-language settings, existing KD work focuses on model compression, transferring knowledge from a large VLM to a smaller one.
In contrast, we distil from a text-only LM into a multimodal VLM, to recover linguistic capability lost during fine-tuning.

\paragraph{KV-Cache Sharing across Model Components.}
Sharing key-value representations between model components lets one model attend to another's context without re-encoding.
Speculative decoding \citep{leviathan2023fast} lets a verifier extend the draft model's KV cache.
In encoder-decoder architectures, cross-attention over the encoder's KV cache is the standard mechanism for the decoder to access input representations \citep{Vaswani+2017}.
T5Gemma 2 \citep{zhang2025t5gemma2seeingreading} concatenates encoder and decoder KV representations into a single attention module with shared parameters.
UniFusion \citep{li2025unifusionvisionlanguagemodelunified} explores layerwise KV fusion between a VLM encoder and a diffusion decoder, though they find naive concatenation without dedicated projections causes feature misalignment.
We extend the student VLM's KV cache to a frozen LM during training, allowing it to attend to visual context without modifying its parameters, and discard it after training.

\section{\fullmethod}

\fullmethod is a training-time framework that combines architectural modification and selective distillation to recover the native linguistic capability of a VLM. The required components are a VLM and its instruction-tuned LM backbone prior to multimodal alignment, such as LLaVA and its Vicuna backbone \citep{liu2023llava}. The VLM serves as the student and the frozen LM acts as the teacher whose linguistic capability is meant to be distilled back to the student. After training, the teacher is removed, resulting in a standard VLM with no additional parameters or inference overhead, which we denote as \textsc{LinguDistill} for the remainder of the paper.

\subsection{KV Sharing Architecture}
\begin{wrapfigure}[12]{r}{0.37\linewidth}
\vspace{-18pt}
\centering
\includegraphics[width=\linewidth]{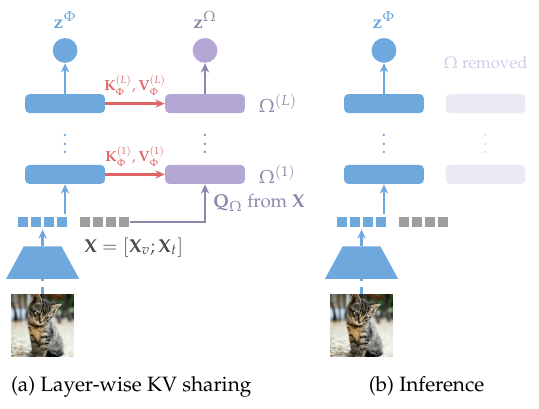}
\captionsetup{font=small,justification=raggedright,singlelinecheck=false}
\caption{KV sharing (a) and inference (b).}
\label{fig:kv-arch}
\vspace{-16pt}
\end{wrapfigure}

To enable frozen LM to have multimodal understanding, we use the VLM itself as the bridge. By taking inspiration from UniFusion's  layer-wise attention fusion \citep{li2025unifusionvisionlanguagemodelunified}, we design layer-wise KV cache sharing (Figure~\ref{fig:kv-arch}). This requires no extra modules, because the teacher and student share the same LM architecture. This allows the teacher to access the same multimodal representation produced by student for decoding context. During training, the student learns to produce multimodal representations that can be interpreted by the frozen LM so it can generate vision-aware supervision signals.

We first process the image using a vision encoder followed by a projector, mapping visual features into the language embedding space. These features are then combined with text tokens to form a multimodal sequence $\mathbf{X} = [\mathbf{X}_v; \mathbf{X}_t]$. This multimodal representation can be obtained either through a standard encoder–projector pipeline or via unified paradigms (e.g., discrete tokenization or codebook-based representations). Ultimately, all cross-modal interaction happens at the decoder level. We denote the student decoder as $\Phi$ and the frozen teacher decoder as $\Omega$. The KV-cache sharing mechanism proceeds as follows:

\begin{enumerate}[leftmargin=*]
    \item The student decoder $\Phi$ processes the full multimodal sequence $\mathbf{X}$ and produces KV caches:
    \begin{equation*}
    \mathbf{K}_\Phi^{(l)}, \mathbf{V}_\Phi^{(l)} = \mathbf{X}\mathbf{W}_{k,\Phi}^{(l)},\; \mathbf{X}\mathbf{W}_{v,\Phi}^{(l)}, \quad l = 1, \dots, L
    \end{equation*}

    \item The teacher decoder $\Omega$ reprocesses the same text prompt to form its query states:
    \begin{equation*}
    \mathbf{Q}_\Omega^{(l)} = \mathbf{X}\mathbf{W}_{q,\Omega}^{(l)}, \quad l = 1, \dots, L
    \end{equation*}

    \item The teacher $\Omega$ directly reuses the student KV cache at every layer:
    \begin{equation*}
    \mathbf{K}_*^{(l)} = \mathbf{K}_\Phi^{(l)}, \quad
    \mathbf{V}_*^{(l)} = \mathbf{V}_\Phi^{(l)}
    \end{equation*}

    \item The teacher attends using its computed queries over the transported student memory:
    \begin{equation*}
    \mathbf{A}^{(l)} = \mathrm{Softmax}\!\left(\frac{\mathbf{Q}_\Omega^{(l)}(\mathbf{K}_\Phi^{(l)})^T}{\sqrt{d_h}} + \mathbf{M}\right)\mathbf{V}_\Phi^{(l)}
    \end{equation*}
\end{enumerate}

This design allows the teacher to attend to the student’s multimodal context via shared KV caches, which conditions it on the multimodal input. The approach is general and extends beyond bimodal interaction as well.

\subsection{Selective Distillation Objective.}
\label{subsec:select-distill}
The student ($\Phi$) is optimized with a typical KD mixture of frozen LM's ($\Omega$) soft distillation and its own hard next-token supervision. Let $\Omega_{\text{pos}}$ denote the set of non-padding positions of teacher models' output whose labels are not ignored. With temperature $T$, the soft distillation loss is

\begin{equation*}
\mathcal{L}_{\mathrm{soft}}
=
\frac{T^{2}}{|\Omega_{\text{pos}}|}
\sum_{(b,t)\in\Omega_{\text{pos}}}
\mathrm{KL}\!\left(
\mathrm{softmax}(\mathbf{z}^{\Omega}_{b,t}/T)\,\|\,\mathrm{softmax}(\mathbf{z}^{\Phi}_{b,t}/T)
\right)
\end{equation*}

We also retain the standard hard-label objective on the student:

\begin{equation*}
\mathcal{L}_{\mathrm{hard}}
=
\frac{1}{|\Omega_{\text{pos}}|}
\sum_{(b,t)\in\Omega_{\text{pos}}}
\mathrm{CE}(\mathbf{z}^{\Phi}_{b,t}, y_{b,t})
\end{equation*}

Instead of applying uniform distillation, \fullmethod uses data-dependent weighting, where distillation is primarily applied to language-intensive data. The final objective is

\begin{equation*}
\mathcal{L}
=
\frac{1}{|\Omega_{\text{pos}}|}
\sum_{(b,t)\in\Omega_{\text{pos}}}
\alpha(d_b)\,\mathcal{L}^{(b,t)}_{\mathrm{soft}}
+
\left(1-\alpha(d_b)\right)\,\mathcal{L}^{(b,t)}_{\mathrm{hard}}
\end{equation*}

where $d_b$ denotes the data source of example $b$.

This objective allows the teacher's supervision signal to be sourced more from language-relevant data, while relying more on hard supervision for visual-heavy sources. As a result, \fullmethod restores linguistic capability by applying distillation only on language-capability recovering data, where the LM provides strong supervision, while preserving the original structure of the VLM on visual-heavy or general multimodal data.

\section{Setup}

\subsection{Dataset Design}
\label{sec:data}


All experiments use The Cauldron \citep{laurencon2024matters}, a collection of 50 VL instruction-tuning datasets. From there, we retain 17 core sources covering a balanced mix of VQA, OCR/document, knowledge, and domain-specific tasks. This selection reduces the total dataset size to under 600k examples while preserving the most relevant training signals. From these 17 sources, we further define a subset of 8 language-intensive VL datasets for data ablation. We refer to these two settings as \texttt{full} (17 sources) and \texttt{lang-subset} (8 sources) and the summary table can be observed in Table \ref{tab:sources}. The full set serves as the baseline, simulating continued multimodal fine-tuning and enabling selective distillation via source identifiers. In contrast, the lang-subset is used to test whether restricting training to language-heavy data alone can recover linguistic capability. All samples contain 1 image, with a maximum sequence length of 1024 tokens, with rationales detailed in Appendix~\ref{app:seqlen}.




\begin{table}[h]
\centering
\caption{Training source categorisation. Language-heavy sources receive the full KD signal ($\alpha > 0$); OCR/document sources receive CE only ($\alpha = 0$) in selective distillation variants.}
\label{tab:sources}
\setlength{\tabcolsep}{5pt}
\small
\begin{tabular}{llcc}
\toprule
\textbf{Category} & \textbf{Sources} & \textbf{KD signal} & \textbf{Reasoning} \\
\midrule
Language-heavy (8) &
  \begin{tabular}[c]{@{}l@{}}A-OKVQA, FigureQA, IconQA,\\RobustQA, ScienceQA,\\Visual7W, VQAv2, VSR\end{tabular}
  & $\alpha > 0$ &
  \begin{tabular}[c]{@{}l@{}}Distill teacher’s linguistic\\ priors on reasoning\\and knowledge tasks\end{tabular} \\
\addlinespace
OCR/doc-heavy (9) &
  \begin{tabular}[c]{@{}l@{}}Chart2Text, ChartQA, DocVQA,\\InfographicVQA, OCR-VQA,\\TextCaps, TextVQA,\\VisText, VisualMRC\end{tabular}
  & $\alpha = 0$ &
  \begin{tabular}[c]{@{}l@{}}Preserve student’s native\\visual and OCR grounding\\ability\end{tabular} \\
\bottomrule
\end{tabular}
\end{table}

\subsection{Models \& Optimization}
\label{sec:data}

We use the instruction-tuned nanoVLM-460M-8k model \citep{wiedmann2025nanovlm} as our VLM, along with its instruction-tuned LM backbone, SmolLM-360M-Instruct \citep{allal2025smollm2}, as the teacher. The VLM serves as the student, with vision encoder being frozen, while the LM remains frozen during training. Both models are capped at a maximum input length of 1024 tokens, aligned with the dataset design. For training, we use the selective distillation objective described in Sec.~\ref{subsec:select-distill}. All variants are trained with the same learning rate at 1e-4, \texttt{bf16} precision, and token budget at $\sim$380M tokens to 4000 steps using cosine LR scheduler. Expanded training details are provided in the Appendix \ref{app:training} and \ref{app:train-pseudo}.

\subsection{Experiment Runs}
\label{sec:variants}

We design 10 run variants which serve as a baseline, method verification, and further data and hyperparameter ablations.
All variants share the same base model, \texttt{lusxvr/nanoVLM-460M-8k} \citep{wiedmann2025nanovlm}, and draw from the data pool explained in Sec. \ref{sec:data}. Table~\ref{tab:variants} summarises the configuration of each variant.

\definecolor{lightgray}{RGB}{240,240,240}
\definecolor{lightyellow}{RGB}{255,248,220}
\definecolor{lightblue}{RGB}{220,235,255}
\definecolor{darkblue}{RGB}{180,210,255}

\begin{table}[h]
\centering
\caption{Experimental variants: baselines (gray), uniform KD (yellow), selective KD (blue).}
\label{tab:variants}
\setlength{\tabcolsep}{6pt}
\small
\begin{tabular}{lcccc}
\toprule
\textbf{Variant} & \textbf{KD} & $\boldsymbol{\alpha}$ & $\boldsymbol{T}$ & \textbf{Notes} \\
\midrule

\multicolumn{5}{l}{\textit{Baselines}} \\

\cellcolor{lightgray}\texttt{nanoVLM} & -- & -- & -- & Pretrained checkpoint \\

\cellcolor{lightgray}\texttt{nanoVLM-full} & $\times$ & -- & -- & Fine-tune on all sources \\

\cellcolor{lightgray}\texttt{nanoVLM-lang} & $\times$ & -- & -- & Fine-tune on language subset \\

\midrule

\multicolumn{5}{l}{\textit{Uniform KD}} \\

\cellcolor{lightyellow}\texttt{distill-full} & $\checkmark$ & 0.5 & 2 & KD on \texttt{full} subset \\

\cellcolor{lightyellow}\texttt{distill-lang} & $\checkmark$ & 0.5 & 2 & KD on \texttt{lang} subset \\

\midrule

\multicolumn{5}{l}{\textit{Selective KD (Ours)}} \\


\cellcolor{lightblue}\texttt{LinguDistill} & $\checkmark$ & 0.7 & 4 & Selective KD on full data \\

\cellcolor{darkblue}\texttt{LinguDistill-highKD} & $\checkmark$ & 0.7 & 4 & Higher Teacher Involvement \\

\cellcolor{darkblue}\texttt{LinguDistill-lowKD} & $\checkmark$ & 0.3 & 2 & Lower Teacher Involvement \\

\bottomrule
\end{tabular}
\end{table}

The variants are organized into three groups. 
\colorbox{lightgray}{\textbf{Baselines}} establish the effect of standard fine-tuning without distillation, comparing the pretrained \basevlm{} checkpoint with full-data fine-tuning and a language-only subset (\texttt{nanoVLM-full}, \texttt{nanoVLM-lang}). These measure how linguistic capability degrades under multimodal adaptation and whether data restriction alone can mitigate it. 
\colorbox{lightyellow}{\textbf{Uniform KD}} introduces distillation from the frozen LM across all data (\texttt{distill-full}), testing whether the teacher can recover linguistic capability. While this improves language-heavy benchmarks, it consistently harms vision-heavy tasks due to misalignment between text-only supervision and visual grounding. 
\colorbox{lightblue}{\textbf{Selective KD (Ours)}} resolves this trade-off by routing the distillation signal based on data type. \texttt{LinguDistill-lang} applies KD only on language-heavy data, while \texttt{LinguDistill-full} retains all sources but suppresses KD on OCR/document tasks, preserving visual alignment while recovering linguistic capability. Additional ablations on temperature and distillation weight confirm that the gains arise from selective routing rather than hyperparameter choice.

\subsection{Evaluation}
\label{sec:evaluation}








We evaluate all variants using lmms-eval \citep{zhang2024lmmsevalrealitycheckevaluation} in a zero-shot setting with batch size 1. Rather than reporting a single aggregate score, we organize benchmarks into three groups that reflect distinct capability axes, so we can see where linguistic recovery helps, where it hurts, and where effects are mixed. Framework identifier specific detail can be observed in Appendix \ref{app:lmms-eval}.

\textbf{Language and knowledge-heavy benchmarks} comprise ARC Easy and Challenge \citep{clark2018arc}, HellaSwag \citep{zellers2019hellaswag}, ScienceQA \citep{lu2022scienceqa}, and COCO 2017 captioning \citep{chen2015microsoftcococaptionsdata}. These tasks are primarily solved through linguistic reasoning and world knowledge, with visual content playing a secondary role. We expect \fullmethod to show the gains here, as the distilled linguistic priors of the original LM backbone are most directly beneficial for this suite.

\textbf{Document, OCR, and vision-specific benchmarks} comprise DocVQA \citep{mathew2021docvqa}, InfographicVQA \citep{mathew2022infographicvqa}, OCRBench \citep{liu2024ocrbench}, and RealWorldQA \citep{xai2024realworldqa}. These tasks demand fine-grained visual perception and are less reliant on linguistic priors. We expect some degradation here relative to standard fine-tuning, as distillation pressure toward language capability may reduce understanding on vision-heavy signals.

\textbf{General multimodal benchmarks} comprise MMMU \citep{yue2024mmmumassivemultidisciplinemultimodal}, MMStar \citep{mmstar}, MME perception and cognition \citep{fu2023mme}, and AI2D \citep{kembhavi2016ai2d}. These benchmarks blend visual understanding with knowledge and reasoning, making them a useful middle ground for assessing whether \fullmethod preserves overall multimodal competence while recovering linguistic capability.









\section{Results}
\label{sec:results}

\definecolor{redDark}{HTML}{C0392B}
\definecolor{redMid}{HTML}{E07060}
\definecolor{grnDark}{HTML}{1E7D34}
\definecolor{neutralGray}{HTML}{F8F8F8}

Table~\ref{tab:summary-results} summarises category averages for every variant, and Appendix~\ref{app:full-results} lists all benchmark scores.
All variants start from the \texttt{lusxvr/nanoVLM-460M-8k} checkpoint and are evaluated at step 4000. We report changes relative to \basevlm{} because it shares every training condition with our variants (data, steps, token budget and learning rate) except the distillation signal, so the difference isolates the effect of distillation. Since \basevlm{} has itself lost ground, we also state how each variant compares with the original \vanillavlm{} checkpoint, which measures how much of the lost capability is actually recovered.

\providecommand{\sincr}[1]{\textcolor{green!60!black}{#1}}
\providecommand{\sdecr}[1]{\textcolor{red!70!black}{#1}}
\providecommand{\sneut}[1]{\textcolor{gray}{#1}}
\begin{table}[t]
\centering
\caption{Category averages across variants (full per-benchmark scores in Appendix~\ref{app:full-results}). \textit{Language-heavy} averages AI2D, COCO, ScienceQA, ARC-Easy, ARC-Challenge and HellaSwag; \textit{Text-only} is its ARC/HellaSwag subset, which has no visual input. The original checkpoint (grey) is shown for reference; bold marks the best trained variant per column. Small coloured percentages give the relative change versus the original checkpoint.}
\label{tab:summary-results}
\setlength{\tabcolsep}{4pt}
\small
\begin{tabular}{l cc c cc}
\toprule
& \multicolumn{2}{c}{\textbf{Language}} & \textbf{Doc \& OCR} & \multicolumn{2}{c}{\textbf{General multimodal}} \\
\cmidrule(lr){2-3}\cmidrule(lr){4-4}\cmidrule(lr){5-6}
\textbf{Variant} & Language-heavy & Text-only & Avg & MME & Others \\
\midrule
{\color{gray}\vanillavlm{} (original)} & {\color{gray}0.527} & {\color{gray}0.444} & {\color{gray}0.629} & {\color{gray}780.5} & {\color{gray}0.401} \\
\midrule
\rowcolor{lightgray} \texttt{nanoVLM-full} & 0.471\,{\tiny \textcolor{lossR!57!fadeG}{(-10.6\%)}} & 0.382\,{\tiny \textcolor{lossR!67!fadeG}{(-14.0\%)}} & \textbf{0.592}\,{\tiny \textcolor{lossR!43!fadeG}{(-5.9\%)}} & 628.0\,{\tiny \textcolor{lossR!84!fadeG}{(-19.5\%)}} & \textbf{0.390}\,{\tiny \textcolor{lossR!33!fadeG}{(-2.7\%)}} \\
\rowcolor{lightgray} \texttt{nanoVLM-lang} & 0.481\,{\tiny \textcolor{lossR!51!fadeG}{(-8.7\%)}} & 0.384\,{\tiny \textcolor{lossR!66!fadeG}{(-13.5\%)}} & 0.587\,{\tiny \textcolor{lossR!45!fadeG}{(-6.7\%)}} & 665.5\,{\tiny \textcolor{lossR!69!fadeG}{(-14.7\%)}} & 0.383\,{\tiny \textcolor{lossR!38!fadeG}{(-4.5\%)}} \\
\rowcolor{lightyellow} \texttt{distill-full} & 0.524\,{\tiny \textcolor{lossR!27!fadeG}{(-0.6\%)}} & 0.388\,{\tiny \textcolor{lossR!63!fadeG}{(-12.6\%)}} & 0.461\,{\tiny \textcolor{lossR!100!fadeG}{(-26.7\%)}} & 643.5\,{\tiny \textcolor{lossR!78!fadeG}{(-17.6\%)}} & 0.361\,{\tiny \textcolor{lossR!55!fadeG}{(-10.0\%)}} \\
\rowcolor{lightyellow} \texttt{distill-lang} & 0.540\,{\tiny \textcolor{gainG!32!fadeG}{(+2.5\%)}} & 0.424\,{\tiny \textcolor{lossR!38!fadeG}{(-4.5\%)}} & 0.512\,{\tiny \textcolor{lossR!81!fadeG}{(-18.6\%)}} & 708.5\,{\tiny \textcolor{lossR!53!fadeG}{(-9.2\%)}} & 0.379\,{\tiny \textcolor{lossR!42!fadeG}{(-5.5\%)}} \\
\rowcolor{lightblue} \adaptdistill{} & \textbf{0.564}\,{\tiny \textcolor{gainG!46!fadeG}{(+7.0\%)}} & \textbf{0.444}\,{\tiny \sneut{($\pm$0.0\%)}} & 0.557\,{\tiny \textcolor{lossR!59!fadeG}{(-11.4\%)}} & \textbf{736.5}\,{\tiny \textcolor{lossR!42!fadeG}{(-5.6\%)}} & 0.374\,{\tiny \textcolor{lossR!45!fadeG}{(-6.7\%)}} \\
\bottomrule
\end{tabular}

\vspace{4pt}
\begin{minipage}{0.92\linewidth}
\footnotesize\textit{Reference points.} All trained variants share data, steps and token budget with \texttt{nanoVLM-full}; it is therefore the baseline that isolates the distillation signal. The original checkpoint marks the level before further multimodal fine-tuning, i.e.\ the target for recovery, and percentages in this table are relative to it.
\end{minipage}
\end{table}

\subsection{RQ1: Data selection cannot undo language erosion}

\paragraph{Standard fine-tuning degrades language capability.}
Comparing \vanillavlm{} and \basevlm{} after 4000 training steps, we observe consistent regression on language-intensive and text-only benchmarks.
COCO captioning drops from 0.800 to 0.673 ({\color{lossR!73!fadeG}$-15.9\%$}), MME cognition from 302 to 229 ({\color{lossR!98!fadeG}$-24.2\%$}), and HellaSwag from 0.405 to 0.326 ({\color{lossR!84!fadeG}$-19.5\%$}).
ARC Easy and ARC Challenge also drop by {\color{lossR!57!fadeG}$-10.7\%$} and {\color{lossR!65!fadeG}$-13.4\%$} respectively.
These are benchmarks where generating or understanding language is the primary challenge.
Meanwhile, DocVQA and MMStar ($-0.3\%$) are barely affected. The degradation is therefore concentrated on linguistic capability rather than visual understanding.

\definecolor{redDark}{HTML}{C0392B}
\definecolor{redMid}{HTML}{E07060}
\definecolor{grnDark}{HTML}{1E7D34}
\definecolor{neutralGray}{HTML}{F8F8F8}

\paragraph{Training on language-heavy data alone does not help.}
\selectvlm{} restricts training to the 8 language-heavy sources, removing all OCR and document data.
This partially improves COCO ({\color{gainG!56!fadeG}$+10.3\%$} relative to \basevlm{}) but shows inconsistent gains on reasoning benchmarks.
HellaSwag improves only marginally to 0.329 ({\color{gainG!28!fadeG}$+0.9\%$}), ARC Easy to 0.542 ({\color{gainG!26!fadeG}$+0.4\%$}), while ARC Challenge remains limited at 0.281 ({\color{gainG!27!fadeG}$+0.7\%$}).
MME cognition improves to 261 ({\color{gainG!67!fadeG}$+14.0\%$}), indicating partial recovery.
However, we also observe slight regressions on AI2D ({\color{lossR!27!fadeG}$-0.7\%$}) and ScienceQA ({\color{lossR!31!fadeG}$-1.9\%$}), which are themselves language- and knowledge-intensive benchmarks.
These degradations are counterintuitive, as such tasks should benefit from restricting training to language-heavy data, and suggest that the underlying representation drift is not resolved by data filtering alone.
Filtering reduces some degradation but does not correct the backbone drift from multimodal fine-tuning.

\subsection{RQ2: A frozen text-only teacher restores language}

\paragraph{Uniform distillation recovers language but damages vision tasks.}
\distill{} applies the KD signal uniformly across all sources with $\alpha{=}0.5$ and $T{=}2$.
Language benchmarks improve clearly over the fine-tuning baseline: COCO rises to 0.843 ({\color{gainG!100!fadeG}$+25.3\%$}), AI2D to 0.490 ({\color{gainG!78!fadeG}$+17.8\%$}), and ScienceQA to 0.650 ({\color{gainG!54!fadeG}$+9.8\%$}).
Text-only reasoning benchmarks also improve modestly: HellaSwag reaches 0.345 ({\color{gainG!42!fadeG}$+5.8\%$}), ARC Easy 0.548 ({\color{gainG!30!fadeG}$+1.5\%$}), while ARC Challenge remains slightly degraded at 0.270 ({\color{lossR!35!fadeG}$-3.2\%$}).
However, OCR and document tasks are severely damaged: OCRBench drops from 0.726 to 0.452 ({\color{lossR!100!fadeG}$-37.7\%$}) and DocVQA from 0.767 to 0.640 ({\color{lossR!75!fadeG}$-16.6\%$}).
This signals that supervision from a pure language teacher, even with KV-cache sharing, misguides the student on pixel-level tasks lacking visual grounding.
We conclude that uniform distillation over-regularizes the model toward language behavior, suppressing modality-specific signals and leading to task-specific degradation.

\paragraph{Selective distillation recovers language fully.}
\adaptdistill{} trains on all 17 sources (\texttt{full} subset) but applies $\alpha{=}0$ to the 9 OCR/document sources, selectively absorbing KD signal only for language-heavy examples.
This results in the strongest improvements across all language benchmarks: ScienceQA reaches 0.676 ({\color{gainG!68!fadeG}$+14.2\%$}), AI2D 0.507 ({\color{gainG!91!fadeG}$+21.9\%$}), COCO 0.866 ({\color{gainG!100!fadeG}$+28.7\%$}), and MME cognition 300 ({\color{gainG!100!fadeG}$+31.0\%$}).
Text-only reasoning benchmarks also show consistent gains: HellaSwag improves to 0.394 ({\color{gainG!88!fadeG}$+20.9\%$}), ARC Easy to 0.621 ({\color{gainG!70!fadeG}$+15.0\%$}), and ARC Challenge to 0.318 ({\color{gainG!67!fadeG}$+14.0\%$}).
These benchmarks have no visual component and directly reflect linguistic capability.

\subsection{RQ3: Selective routing contains the visual cost}

Vision task degradation is much smaller than under uniform distillation: DocVQA drops only {\color{lossR!36!fadeG}$-3.5\%$} (vs.\ {\color{lossR!75!fadeG}$-16.6\%$}), while OCRBench drops {\color{lossR!77!fadeG}$-17.4\%$} (vs.\ {\color{lossR!100!fadeG}$-37.7\%$}).
InfographicVQA instead improves to 0.330 ({\color{gainG!76!fadeG}$+17.0\%$}), indicating better balance between modalities.
We further analyze OCRBench degradation and find that these tasks rely on visual grounding signals unavailable to the text-only teacher (Appendix~\ref{app:ocr}).
MMMU remains degraded at {\color{lossR!51!fadeG}$-8.7\%$} across distillation variants; we attribute this to the single-image, 1024-token training constraint rather than the distillation objective itself (Appendix~\ref{app:seqlen}).

Measured against the original checkpoint, \adaptdistill{} matches \vanillavlm{} on text-only reasoning, exceeds it on COCO, ScienceQA and AI2D, and remains below it on the document and OCR benchmarks, MME and MMMU.
Overall, selective distillation gives the best trade-off: language and text-only reasoning return to, or exceed, the original level, at a fraction of the document and OCR cost of uniform distillation.

\section{Discussion}
\label{sec:discussion}


\subsection{What Does Distillation Recover?}
\label{sec:distill-transfer}

We compare \adaptdistill{} and \basevlm{} one by one to see what the teacher transfers. On ScienceQA, \adaptdistill{} fixes 854 examples that \basevlm{} gets wrong, while losing only 498 that \basevlm{} gets right (+8.39 percentage points net).
All of \basevlm{}'s mistakes are actual wrong answers, not formatting issues.
The gains cover natural science (443), language science (215), and social science (196), spanning physics, biology, language arts, and commonsense reasoning.
The frozen teacher still knows things that the fine-tuned student has forgotten, and \fullmethod brings the knowledge back.

On AI2D, \adaptdistill{} wins 645 examples and loses 362 (+9.16 percentage points), mostly in food web reasoning and biology diagrams.
These tasks need the model to understand how things relate to each other, not just read labels.
InfoVQA follows the same pattern.
Even though it is an OCR-heavy benchmark, \adaptdistill{} improves by +4.70 percentage points (703 wins vs.\ 432 losses) because the questions ask about what the infographic means, not just what it says.
Distillation therefore helps whenever a task needs knowledge from the frozen teacher, whether or not it also involves reading text.

\subsection{Effect on Visual Text Tasks}
\label{sec:ocr-effect}

We break OCRBench into its ten sub-tasks (Appendix~\ref{app:ocr}) to see where the drops come from.
Basic text recognition barely changes ($-4$/150 points). The model can still read printed text.
Most of the drop is in Key Information Extraction ($-63$) and Doc-oriented VQA ($-26$), which together make up 71\% of the total loss.
Both tasks require the model to pick the right field from a structured document like a receipt or form, where many text fields sit close together. A stronger language prior makes the model worse at picking the right spot.

We look at 172 examples where \basevlm{} gets the answer right but \adaptdistill{} does not, and find three failure patterns.
First, the model corrects nonsense strings into real words (e.g., \texttt{PEAEC} becomes \textit{peace}), which fails when the text is not meant to be a real word.
Second, the model reads from the right document but the wrong spot, returning a valid answer from a nearby field.
Third, the model paraphrases instead of copying, rounding numbers and shortening names.
All three come down to the same thing. The teacher's language signal makes the model better at understanding text but worse at copying exactly what it sees. This only shows up in document extraction tasks and does not spread to other benchmarks.

\subsection{Ablations}

\paragraph{Data Subset} Data selection alone does not recover language. Without distillation, training on the \texttt{lang} subset gives almost the same scores as the \texttt{full} subset (Table~\ref{tab:main-results}). With uniform distillation, the \texttt{lang} subset helps, mostly because it removes the OCR and document data on which the text-only teacher gives misleading supervision. Recovery therefore depends on the training objective, not on the data mix.

\paragraph{Selective Distillation} We attribute \fullmethod's success to selective distillation. As observed when comparing \colorbox{lightyellow}{Uniform KD} and \colorbox{lightblue}{\fullmethod} in Table \ref{tab:main-results}, selective distillation keeps more OCR capability and also improves text-only benchmarks. This works because the teacher signal is applied only where it is reliable (language-heavy data), while being suppressed on vision-heavy tasks, preventing misleading supervision and preserving the student’s visual grounding.

\begin{figure}[t]
    \centering
    \begin{subfigure}[t]{0.49\linewidth}
        \centering
        \includegraphics[width=\linewidth]{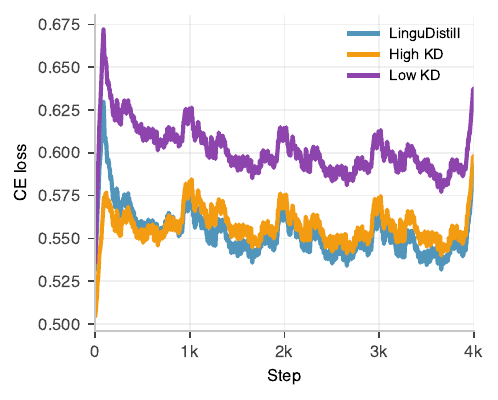}
        \caption{Hard loss across KD-strength variants.}
        \label{fig:discussion-losses-compare}
    \end{subfigure}\hfill
    \begin{subfigure}[t]{0.49\linewidth}
        \centering
        \includegraphics[width=\linewidth]{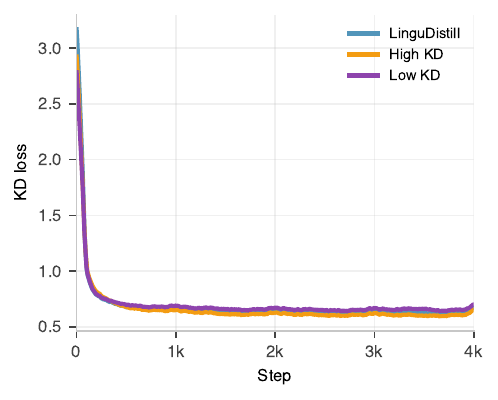}
        \caption{Soft loss across KD-strength variants.}
        \label{fig:discussion-losses-decomp}
    \end{subfigure}
    \caption{Training-loss analysis for the three selective distillation variants. Left: the CE term is lowest for the main \adaptdistill{} setting, with \textit{High KD} slightly above it and \textit{Low KD} clearly worse. Right: the Soft loss uses the combined logged objective, which mixes the teacher-weighted distillation signal with the VLM loss in a balanced manner.}
    \label{fig:discussion-losses}
\end{figure}

\paragraph{Distillation Hyperparams} Figure~\ref{fig:discussion-losses} shows the optimization behavior behind this trade-off.
In Figure~\ref{fig:discussion-losses-compare}, we compare the CE term of our three selective distillation variants: the main \adaptdistill{} setting, a stronger KD configuration (\textit{High KD}), and a weaker one (\textit{Low KD}).
The main \adaptdistill{} setting reaches the lowest CE curve overall, while \textit{High KD} stays slightly above it and \textit{Low KD} remains the highest throughout training.
This indicates that pushing the teacher signal too hard hurts the student's direct supervision, but weakening the KD term too much is also suboptimal.
In Figure~\ref{fig:discussion-losses-decomp}, we compare the corresponding KD losses.
\textit{High KD} gains the lowest soft loss, \textit{Low KD} remains the highest soft loss, and the main \adaptdistill{} setting sits between them.
The best benchmark trade-off comes from the middle setting, where we balance learning through soft loss.

\section{Conclusion}
\label{sec:conclusion}
We study the degradation of linguistic capability in VLMs caused by multimodal adaptation, and show that this loss is not undone by standard fine-tuning or by restricting training to language-heavy data (RQ1). We propose \fullmethod, an adapter-free distillation framework in which the frozen original LM supervises the VLM through layer-wise KV-cache sharing. The teacher recovers language and knowledge performance, including on text-only benchmarks with no visual input (RQ2). Routing the distillation signal by data source keeps the cost on document and OCR tasks far below that of uniform distillation (RQ3). Because the teacher is discarded after training, the resulting model has no extra parameters or inference cost.

\paragraph{Limitations.}
All experiments use a single model family at the 460M scale, with one training run per variant evaluated at a fixed step; confirming the trend at larger scales and across seeds is left to future work. Training uses single-image samples capped at 1024 tokens, which we believe explains the residual MMMU regression (Appendix~\ref{app:seqlen}). Document extraction tasks such as key information extraction remain below the fine-tuning baseline, because a stronger language prior trades exact copying for fluent correction (Section~\ref{sec:ocr-effect}). Finally, the method requires access to the original LM checkpoint during training.

\newpage

\bibliography{colm2026_conference}
\bibliographystyle{colm2026_conference}

\newpage
\appendix
\section{Full Benchmark Results}
\label{app:full-results}
Table~\ref{tab:main-results} reports every benchmark score behind the category averages in Table~\ref{tab:summary-results}. Coloured percentages denote relative change versus \texttt{nanoVLM-full}.

\definecolor{avgGray}{HTML}{E0E0E0}

\newcommand{\incr}[1]{\textcolor{green!60!black}{#1}}
\newcommand{\decr}[1]{\textcolor{red!70!black}{#1}}
\newcommand{\fmt}[2]{#1\hspace{2pt}{\tiny #2}}

\begin{table}[t]
\centering
\caption{
Benchmark results across variants. Values are absolute scores; {\tiny colored percentages} denote change vs.\ \texttt{nanoVLM-full}.
Average rows are manually color-scaled for perceptual consistency: small changes are subtle, larger changes progressively emphasized.
Selective KD (\texttt{LinguDistill}) gives the strongest gains with controlled perception loss.
}
\setlength{\tabcolsep}{3pt}
\scriptsize
\begin{tabular}{l c cc cc c}
\toprule
& & 
\multicolumn{2}{c}{\makecell{\textbf{Fine-tuning}}} 
& 
\multicolumn{2}{c}{\makecell{\textbf{Uniform KD} \\ (Ours w/o Task Weighting)}} 
& 
\multicolumn{1}{c}{\makecell{\textbf{Selective KD} \\ (Ours)}} \\
\cmidrule(lr){3-4}\cmidrule(lr){5-6}\cmidrule(lr){7-7}
\textbf{Task} & \vanillavlm{} &
\cellcolor{lightgray}\texttt{nanoVLM-full} &
\cellcolor{lightgray}\texttt{nanoVLM-lang} &
\cellcolor{lightyellow}\texttt{distill-full} &
\cellcolor{lightyellow}\texttt{distill-lang} &
\cellcolor{lightblue}\texttt{LinguDistill} \\
\midrule

\multicolumn{7}{l}{\textit{Language-heavy}} \\
\midrule
AI2D & {\color{gray}0.440} & 0.416 & \fmt{0.413}{\textcolor{lossR!27!fadeG}{(-0.7\%)}} & \fmt{0.490}{\textcolor{gainG!78!fadeG}{(+17.8\%)}} & \fmt{0.502}{\textcolor{gainG!87!fadeG}{(+20.7\%)}} & \fmt{0.507}{\textcolor{gainG!91!fadeG}{(+21.9\%)}} \\
COCO & {\color{gray}0.800} & 0.673 & \fmt{0.742}{\textcolor{gainG!56!fadeG}{(+10.3\%)}} & \fmt{0.843}{\textcolor{gainG!100!fadeG}{(+25.3\%)}} & \fmt{0.787}{\textcolor{gainG!76!fadeG}{(+16.9\%)}} & \fmt{0.866}{\textcolor{gainG!100!fadeG}{(+28.7\%)}} \\
ScienceQA & {\color{gray}0.590} & 0.592 & \fmt{0.581}{\textcolor{lossR!31!fadeG}{(-1.9\%)}} & \fmt{0.650}{\textcolor{gainG!54!fadeG}{(+9.8\%)}} & \fmt{0.679}{\textcolor{gainG!69!fadeG}{(+14.7\%)}} & \fmt{0.676}{\textcolor{gainG!68!fadeG}{(+14.2\%)}} \\
ARC Easy & {\color{gray}0.605} & 0.540 & \fmt{0.542}{\textcolor{gainG!26!fadeG}{(+0.4\%)}} & \fmt{0.548}{\textcolor{gainG!30!fadeG}{(+1.5\%)}} & \fmt{0.601}{\textcolor{gainG!59!fadeG}{(+11.3\%)}} & \fmt{0.621}{\textcolor{gainG!70!fadeG}{(+15.0\%)}} \\
ARC Challenge & {\color{gray}0.322} & 0.279 & \fmt{0.281}{\textcolor{gainG!27!fadeG}{(+0.7\%)}} & \fmt{0.270}{\textcolor{lossR!35!fadeG}{(-3.2\%)}} & \fmt{0.296}{\textcolor{gainG!43!fadeG}{(+6.1\%)}} & \fmt{0.318}{\textcolor{gainG!67!fadeG}{(+14.0\%)}} \\
HellaSwag & {\color{gray}0.405} & 0.326 & \fmt{0.329}{\textcolor{gainG!28!fadeG}{(+0.9\%)}} & \fmt{0.345}{\textcolor{gainG!42!fadeG}{(+5.8\%)}} & \fmt{0.376}{\textcolor{gainG!71!fadeG}{(+15.3\%)}} & \fmt{0.394}{\textcolor{gainG!88!fadeG}{(+20.9\%)}} \\

\rowcolor{avgGray}
\textbf{Average} &
{\color{gray}\textbf{0.527}} &
\textbf{0.471} &
\cellcolor{green!7}{\fmt{\textbf{0.481}}{\textbf{\textcolor{gainG!31!fadeG}{(+2.1\%)}}}} &
\cellcolor{green!22}{\fmt{\textbf{0.524}}{\textbf{\textcolor{gainG!59!fadeG}{(+11.3\%)}}}} &
\cellcolor{green!28}{\fmt{\textbf{0.540}}{\textbf{\textcolor{gainG!69!fadeG}{(+14.6\%)}}}} &
\cellcolor{green!55}{\fmt{\textbf{0.564}}{\textbf{\textcolor{gainG!84!fadeG}{(+19.7\%)}}}} \\

\midrule
\multicolumn{7}{l}{\textit{Document \& OCR}} \\
\midrule
DocVQA & {\color{gray}0.769} & 0.767 & \fmt{0.766}{\textcolor{lossR!25!fadeG}{(-0.1\%)}} & \fmt{0.640}{\textcolor{lossR!75!fadeG}{(-16.6\%)}} & \fmt{0.709}{\textcolor{lossR!48!fadeG}{(-7.6\%)}} & \fmt{0.740}{\textcolor{lossR!36!fadeG}{(-3.5\%)}} \\
InfographicVQA & {\color{gray}0.357} & 0.282 & \fmt{0.272}{\textcolor{lossR!36!fadeG}{(-3.5\%)}} & \fmt{0.290}{\textcolor{gainG!33!fadeG}{(+2.8\%)}} & \fmt{0.318}{\textcolor{gainG!63!fadeG}{(+12.8\%)}} & \fmt{0.330}{\textcolor{gainG!76!fadeG}{(+17.0\%)}} \\
OCRBench & {\color{gray}0.760} & 0.726 & \fmt{0.722}{\textcolor{lossR!27!fadeG}{(-0.6\%)}} & \fmt{0.452}{\textcolor{lossR!100!fadeG}{(-37.7\%)}} & \fmt{0.510}{\textcolor{lossR!100!fadeG}{(-29.7\%)}} & \fmt{0.600}{\textcolor{lossR!77!fadeG}{(-17.4\%)}} \\

\rowcolor{avgGray}
\textbf{Average} &
{\color{gray}\textbf{0.629}} &
\textbf{0.592} &
\cellcolor{red!6}{\fmt{\textbf{0.587}}{\textbf{\textcolor{lossR!27!fadeG}{(-0.8\%)}}}} &
\cellcolor{red!35}{\fmt{\textbf{0.461}}{\textbf{\textcolor{lossR!91!fadeG}{(-22.1\%)}}}} &
\cellcolor{red!25}{\fmt{\textbf{0.512}}{\textbf{\textcolor{lossR!66!fadeG}{(-13.5\%)}}}} &
\cellcolor{red!15}{\fmt{\textbf{0.557}}{\textbf{\textcolor{lossR!43!fadeG}{(-5.9\%)}}}} \\

\midrule
\multicolumn{7}{l}{\textit{General multimodal}} \\
\midrule
MME Cognition & {\color{gray}302} & 229 & \fmt{261}{\textcolor{gainG!67!fadeG}{(+14.0\%)}} & \fmt{241}{\textcolor{gainG!41!fadeG}{(+5.2\%)}} & \fmt{308}{\textcolor{gainG!100!fadeG}{(+34.5\%)}} & \fmt{300}{\textcolor{gainG!100!fadeG}{(+31.0\%)}} \\
MME Perception & {\color{gray}1259} & 1027 & \fmt{1070}{\textcolor{gainG!38!fadeG}{(+4.2\%)}} & \fmt{1046}{\textcolor{gainG!30!fadeG}{(+1.8\%)}} & \fmt{1109}{\textcolor{gainG!49!fadeG}{(+8.0\%)}} & \fmt{1173}{\textcolor{gainG!68!fadeG}{(+14.2\%)}} \\

\rowcolor{avgGray}
\textbf{MME Avg} &
{\color{gray}\textbf{780.5}} &
\textbf{628.0} &
\cellcolor{green!18}{\fmt{\textbf{665.5}}{\textbf{\textcolor{gainG!43!fadeG}{(+6.0\%)}}}} &
\cellcolor{green!12}{\fmt{\textbf{643.5}}{\textbf{\textcolor{gainG!32!fadeG}{(+2.5\%)}}}} &
\cellcolor{green!30}{\fmt{\textbf{708.5}}{\textbf{\textcolor{gainG!63!fadeG}{(+12.8\%)}}}} &
\cellcolor{green!50}{\fmt{\textbf{736.5}}{\textbf{\textcolor{gainG!77!fadeG}{(+17.3\%)}}}} \\

RealWorldQA & {\color{gray}0.523} & 0.500 & \fmt{0.488}{\textcolor{lossR!32!fadeG}{(-2.4\%)}} & \fmt{0.450}{\textcolor{lossR!55!fadeG}{(-10.0\%)}} & \fmt{0.494}{\textcolor{lossR!29!fadeG}{(-1.2\%)}} & \fmt{0.490}{\textcolor{lossR!31!fadeG}{(-2.0\%)}} \\
MMMU & {\color{gray}0.320} & 0.310 & \fmt{0.302}{\textcolor{lossR!33!fadeG}{(-2.6\%)}} & \fmt{0.283}{\textcolor{lossR!51!fadeG}{(-8.7\%)}} & \fmt{0.288}{\textcolor{lossR!46!fadeG}{(-7.1\%)}} & \fmt{0.283}{\textcolor{lossR!51!fadeG}{(-8.7\%)}} \\
MMStar & {\color{gray}0.360} & 0.359 & \fmt{0.358}{\textcolor{lossR!26!fadeG}{(-0.3\%)}} & \fmt{0.350}{\textcolor{lossR!32!fadeG}{(-2.5\%)}} & \fmt{0.355}{\textcolor{lossR!28!fadeG}{(-1.1\%)}} & \fmt{0.349}{\textcolor{lossR!33!fadeG}{(-2.8\%)}} \\

\rowcolor{avgGray}
\textbf{Non-MME Avg} &
{\color{gray}\textbf{0.401}} &
\textbf{0.390} &
\cellcolor{red!7}{\fmt{\textbf{0.383}}{\textbf{\textcolor{lossR!30!fadeG}{(-1.8\%)}}}} &
\cellcolor{red!22}{\fmt{\textbf{0.361}}{\textbf{\textcolor{lossR!47!fadeG}{(-7.4\%)}}}} &
\cellcolor{red!12}{\fmt{\textbf{0.379}}{\textbf{\textcolor{lossR!33!fadeG}{(-2.8\%)}}}} &
\cellcolor{red!15}{\fmt{\textbf{0.374}}{\textbf{\textcolor{lossR!37!fadeG}{(-4.1\%)}}}} \\

\bottomrule
\end{tabular}
\label{tab:main-results}
\end{table}

\section{Training Configuration}
\label{app:training}
\begin{table}[H]
\centering
\caption{Shared training configuration across all variants.}
\label{tab:shared-config}
\small
\begin{tabular}{ll}
\toprule
\textbf{Parameter} & \textbf{Value} \\
\midrule
Vision encoder & SigLIP2-B/16-512 \citep{tong2025siglip2} (frozen) \\
Language decoder & SmolLM2-360M-Instruct \citep{allal2025smollm2} \\
Image tokens & 64 ($4{\times}$ pixel shuffle) \\
Max images per example & 1 \\
Max sequence length & 1024 tokens \\
Effective batch size & 128 \\
Precision & Mixed Precision BF16 \\
Optimizer & AdamW \\
LR (projector) & $1{\times}10^{-4}$ \\
LR (language decoder) & $1{\times}10^{-4}$ \\
LR (vision encoder) & 0 (frozen) \\
LR schedule & Cosine decay \\
Warmup & 3\% of total steps \\
Gradient clipping & Norm 1.0 \\
Dataset & The Cauldron \citep{laurencon2024matters} (filtered) \\
Quality filters & Relevance, image corr., visual dep., formatting $\geq 1$ \\
\bottomrule
\end{tabular}
\end{table}

All experiments use \basevlm{} \citep{wiedmann2025nanovlm} with a total size of 460M parameters, which pairs a SigLIP2-B/16 vision encoder \citep{tong2025siglip2} with a \texttt{SmolLM2-360M-Instruct} language decoder \citep{allal2025smollm2}.
We initialise from the \texttt{lusxvr/nanoVLM-460M-8k} checkpoint and train on a filtered subset of The Cauldron \citep{laurencon2024matters}.
All runs use a single RTX A6000 PRO (96\,GB).

Table~\ref{tab:shared-config} lists hyperparameters shared across all variants.

For distillation variants, the teacher is a frozen copy of the original \texttt{SmolLM2-360M-Instruct} backbone used in \basevlm{}.
The teacher receives zero learning rate throughout training.
All variants are evaluated at step 4000 unless noted otherwise in the main text.

\raggedbottom
\section{Evaluation Subsets}
\label{app:lmms-eval}

We evaluate all models using the \texttt{lmms-eval} framework \citep{zhang2024lmmsevalrealitycheckevaluation} with a standardized zero-shot protocol. All evaluations are performed with a fixed batch size of 1 on a single GPU, and predictions are logged for further analysis.

The evaluation tasks are selected directly from \texttt{lmms-eval} and correspond to the benchmark groups described in Sec.~\ref{sec:evaluation}. Specifically, we use:
\begin{itemize}
    \item \textbf{Language and knowledge-heavy:} \texttt{arc\_easy}, \texttt{arc\_challenge}, \texttt{hellaswag}, \texttt{scienceqa}, \texttt{coco2017\_cap\_val}
    \item \textbf{Document and OCR:} \texttt{ocrbench}, \texttt{docvqa\_val}, \texttt{infovqa\_val}, \texttt{realworldqa}
    \item \textbf{General multimodal:} \texttt{ai2d}, \texttt{mmstar}, \texttt{mme}, \texttt{mmmu\_val}
\end{itemize}
\section{Optimization Pseudocode}
\label{app:train-pseudo}

\begin{figure}[H]
\centering
\small
\begin{minipage}{0.96\textwidth}
\hrule
\vspace{0.4em}
\textbf{Algorithm 1} \ \fullmethod optimization pseudocode
\vspace{0.4em}
\hrule
\vspace{0.5em}
\textbf{Goal:} Optimize the left-side VLM parameters using a detached right-decoder generation built via continuation over a shared KV memory.\\
\textbf{Require:} Minibatch $\mathcal{B}=\{(x_b,I_b,m_b,y_b,d_b)\}_{b=1}^{B}$, image encoder $E_{\mathrm{img}}$, projector $P$, left decoder $D_{\mathrm{left}}$, right decoder $D_{\mathrm{right}}$, left head $H_{\mathrm{left}}$, right head $H_{\mathrm{right}}$, temperature $T$, source weighting rule $\alpha(\cdot)$.\\
\textbf{Ensure:} Scalar loss $\mathcal{L}$ and updated left-side parameters $\theta_{\mathrm{left}}$.\\[0.4em]
\begin{tabular}{@{}r p{0.93\textwidth}@{}}
1 & Initialize $\mathcal{L}_{\mathrm{sum}} \leftarrow 0$ and $N \leftarrow 0$ \\
2 & Freeze $D_{\mathrm{right}}$ and $H_{\mathrm{right}}$ \\
3 & \textbf{for} $b \leftarrow 1$ \textbf{to} $B$ \textbf{do} \\
4 & \hspace{1.5em} $\mathbf{v}_b \leftarrow E_{\mathrm{img}}(I_b)$ \\
5 & \hspace{1.5em} $\mathbf{u}_b \leftarrow P(\mathbf{v}_b)$ \\
6 & \hspace{1.5em} $\tilde{\mathbf{x}}_b \leftarrow \mathrm{concat}(x_b,\mathbf{u}_b)$ \\
7 & \hspace{1.5em} $(\mathbf{h}^{l}_b,\mathrm{KV}^{l}_b) \leftarrow D_{\mathrm{left}}(\tilde{\mathbf{x}}_b,m_b)$ \\
8 & \hspace{1.5em} $\mathbf{z}^{l}_b \leftarrow H_{\mathrm{left}}(\mathbf{h}^{l}_b)$ \\
9 & \hspace{1.5em} $\mathbf{h}^{r}_b \leftarrow D_{\mathrm{right}}(\tilde{\mathbf{x}}_b,m_b;\mathrm{KV}^{l}_b)$ \\
10 & \hspace{1.5em} $\mathbf{z}^{r}_b \leftarrow \mathrm{stopgrad}(H_{\mathrm{right}}(\mathbf{h}^{r}_b))$ \\
11 & \hspace{1.5em} $\alpha_b \leftarrow \alpha(d_b)$ \\
12 & \hspace{1.5em} \textbf{for each} $t$ \textbf{such that} $m_{b,t}=1$ \textbf{and} $y_{b,t}\neq -100$ \textbf{do} \\
13 & \hspace{3em} $p^{l}_{b,t} \leftarrow \mathrm{softmax}(\mathbf{z}^{l}_{b,t}/T)$ \\
14 & \hspace{3em} $p^{r}_{b,t} \leftarrow \mathrm{softmax}(\mathbf{z}^{r}_{b,t}/T)$ \\
15 & \hspace{3em} $\ell_{b,t} \leftarrow \alpha_b T^2 \mathrm{KL}(p^{r}_{b,t}\,\|\,p^{l}_{b,t}) + (1-\alpha_b)\mathrm{CE}(\mathbf{z}^{l}_{b,t},y_{b,t})$ \\
16 & \hspace{3em} $\mathcal{L}_{\mathrm{sum}} \leftarrow \mathcal{L}_{\mathrm{sum}} + \ell_{b,t}$ \\
17 & \hspace{3em} $N \leftarrow N + 1$ \\
18 & \hspace{1.5em} \textbf{end for} \\
19 & \textbf{end for} \\
20 & $\mathcal{L} \leftarrow \mathcal{L}_{\mathrm{sum}} / N$ \\
21 & Update $\theta_{\mathrm{left}}=\{E_{\mathrm{img}},P,D_{\mathrm{left}},H_{\mathrm{left}}\}$ using $\nabla_{\theta_{\mathrm{left}}}\mathcal{L}$ \\
22 & Return $\mathcal{L}$ \\
\end{tabular}
\vspace{0.4em}
\hrule
\end{minipage}
\caption{Pseudo-code for \fullmethod. The image is first mapped into the language subspace, the left decoder absorbs the full multimodal prompt and materializes prompt KV memory, and the frozen right decoder continues from that inherited state to produce detached supervision. The optimization objective combines source-selective distillation with hard next-token supervision, while updating only the left-side parameters.}
\label{fig:lingudistill_algorithm}
\end{figure}

\section{Sequence Length Analysis}
\label{app:seqlen}

We train with a maximum sequence length of 1024 tokens and a single image per example.
This section justifies these choices with token-length statistics from the training data and a discussion of compute constraints.

\paragraph{Training data token distributions.}
Table~\ref{tab:seqlen-train} reports token-length statistics for each training source, computed on a sample of up to 1000 examples per source using the SmolLM2-360M-Instruct tokenizer.
The vast majority of examples fit within 1024 tokens.
Only \texttt{robut\_sqa} (2.4\%) and \texttt{scienceqa} (2.1\%) exceed this threshold, and in both cases the truncation affects fewer than 3\% of examples.
OCR and document sources have mean lengths under 130 tokens, well within the budget.
The overall mean across all sources is under 120 tokens, meaning the 1024-token limit provides ample headroom for most training examples.

\begin{table}[h]
\centering
\caption{Token-length statistics per training source, sampled from The Cauldron and tokenized with SmolLM2-360M-Instruct. ``Cat.'' indicates language-heavy (lang) or OCR/document (ocr) sources. ``\%${>}$1024'' is the fraction of examples exceeding 1024 tokens.}
\label{tab:seqlen-train}
\setlength{\tabcolsep}{5pt}
\small
\begin{tabular}{llrrrrc}
\toprule
\textbf{Source} & \textbf{Cat.} & \textbf{Mean} & \textbf{Med.} & \textbf{P95} & \textbf{Max} & \textbf{\%${>}$1024} \\
\midrule
\texttt{aokvqa} & lang & 62 & 60 & 100 & 116 & 0.0\% \\
\texttt{figureqa} & lang & 212 & 187 & 307 & 321 & 0.0\% \\
\texttt{iconqa} & lang & 37 & 35 & 56 & 664 & 0.0\% \\
\texttt{robut\_sqa} & lang & 180 & 102 & 425 & 1974 & 2.4\% \\
\texttt{scienceqa} & lang & 493 & 118 & 549 & 15171 & 2.1\% \\
\texttt{visual7w} & lang & 251 & 197 & 705 & 972 & 0.0\% \\
\texttt{vqav2} & lang & 79 & 58 & 192 & 673 & 0.0\% \\
\texttt{vsr} & lang & 38 & 26 & 80 & 98 & 0.0\% \\
\addlinespace
\texttt{chart2text} & ocr & 111 & 88 & 225 & 329 & 0.0\% \\
\texttt{chartqa} & ocr & 43 & 35 & 92 & 134 & 0.0\% \\
\texttt{docvqa} & ocr & 107 & 90 & 265 & 357 & 0.0\% \\
\texttt{infographic\_vqa} & ocr & 110 & 97 & 234 & 310 & 0.0\% \\
\texttt{ocrvqa} & ocr & 102 & 102 & 124 & 162 & 0.0\% \\
\texttt{textcaps} & ocr & 22 & 21 & 30 & 35 & 0.0\% \\
\texttt{textvqa} & ocr & 31 & 34 & 48 & 56 & 0.0\% \\
\texttt{vistext} & ocr & 128 & 126 & 182 & 217 & 0.0\% \\
\texttt{visualmrc} & ocr & 99 & 70 & 241 & 292 & 0.0\% \\
\bottomrule
\end{tabular}
\end{table}

\paragraph{Evaluation benchmarks.}
Our evaluation suite spans both short-sequence benchmarks (MME perception and cognition, RealWorldQA, ARC, MMStar) and longer-sequence benchmarks (DocVQA, InfoVQA, OCRBench, MMMU).
The 1024-token training limit does not directly constrain evaluation, since generation uses a separate maximum output length.
However, benchmarks that require long-context document understanding (DocVQA, InfoVQA) may be indirectly affected if the model has not seen sufficiently long sequences during training.
The residual regressions on these benchmarks under selective distillation ($-3.8\%$ on DocVQA, $-7.6\%$ on InfoVQA) are consistent with this explanation.
MMMU additionally requires multi-image reasoning, which our single-image training constraint does not support.

\paragraph{Compute constraints.}
All experiments run on a single RTX A6000 PRO (96\,GB). With BF16 precision, an effective batch size of 128, and the dual-tower distillation architecture (student + frozen teacher), a sequence length of 1024 fits comfortably in memory. Given that fewer than 1\% of training examples exceed 1024 tokens, we prioritize training throughput over marginal coverage. Future work at larger scale can explore extended sequence lengths to improve document-heavy benchmarks.

\section{Detailed OCR Analysis}
\label{app:ocr}

This appendix breaks down the OCRBench results from Section~\ref{sec:discussion}.

\subsection{Sub-task Breakdown}

OCRBench \citep{liu2024ocrbench} has 1{,}000 samples across ten sub-tasks.
We group them by how much the model needs to understand document layout to answer correctly (Table~\ref{tab:ocrbench-subtask}).

\begin{table}[t]
\centering
\caption{OCRBench sub-task breakdown grouped by cognitive demand. \adaptdistill{} vs.\ \basevlm{} (460M, 4k steps). Text recognition (decoding-only) is largely preserved; the regression concentrates in extraction tasks requiring spatial grounding (\textbf{bold}).}
\label{tab:ocrbench-subtask}
\small
\begin{tabular}{llrrr}
\toprule
\textbf{Group} & \textbf{Sub-task} & \textbf{\fullmethod} & \basevlm{} & \textbf{$\Delta$} \\
\midrule
\multirow{3}{*}{\rotatebox[origin=c]{0}{\parbox{2cm}{\centering Text Rec.}}}
  & Regular /50 & 47 & 48 & $-$1 \\
  & Irregular /50 & 40 & 44 & $-$4 \\
  & Artistic /50 & 46 & 45 & +1 \\
\cmidrule{2-5}
  & \emph{Subtotal /150} & \emph{133} & \emph{137} & \emph{$-$4} \\
\midrule
\multirow{4}{*}{\rotatebox[origin=c]{0}{\parbox{2cm}{\centering Handwriting \&\\Non-Semantic}}}
  & Handwriting /50 & 30 & 36 & $-$6 \\
  & Digit String /50 & 38 & 42 & $-$4 \\
  & Non-Semantic /50 & 39 & 46 & $-$7 \\
  & Handwritten Math /100 & 46 & 52 & $-$6 \\
\cmidrule{2-5}
  & \emph{Subtotal /250} & \emph{153} & \emph{176} & \emph{$-$23} \\
\midrule
\multicolumn{2}{l}{Scene Text VQA /200} & 140 & 150 & $-$10 \\
\multicolumn{2}{l}{\textbf{Key Information Extraction /200}} & 84 & 147 & \textbf{$-$63} \\
\multicolumn{2}{l}{\textbf{Doc-oriented VQA /200}} & 90 & 116 & \textbf{$-$26} \\
\midrule
\multicolumn{2}{l}{\textbf{Total /1000}} & 600 & 726 & $-$126 \\
\bottomrule
\end{tabular}
\end{table}

\paragraph{Text Recognition (150 samples).}
This group tests reading printed text in regular, irregular (warped), and artistic (stylised) fonts. The model just needs to read the characters, with no document layout involved.
\adaptdistill{} scores 133/150 vs.\ \basevlm{}'s 137/150, a gap of only 4 points. Regular text drops by one point (47 vs.\ 48) and artistic text improves by one (46 vs.\ 45), showing that the language prior can help with slightly ambiguous text.

\paragraph{Handwriting \& Non-Semantic (250 samples).}
This group covers handwritten words, digit strings, random letter strings (e.g., \texttt{PEAEC}), and math expressions, all of which are visually ambiguous. A language prior can either help or hurt here.
The group drops by $-23$ points (153/250 vs.\ 176/250), with the biggest single drop in non-semantic text ($-7$). This is where the language prior hurts most, since the ground truth is not a real word by design.

\paragraph{Layout-dependent tasks (600 samples).}
\textbf{Scene Text VQA} (200 samples) adds a reasoning step where the model reads text in natural scenes and answers questions about it. \textbf{Key Information Extraction} (200 samples) and \textbf{Doc-oriented VQA} (200 samples) need the model to find the right field in structured documents (receipts, forms, tables) where many text fields sit close together.
Key Information Extraction and Doc-oriented VQA together make up 89 of the 126-point total drop (71\%), even though they are only 40\% of the benchmark. Overall, the three layout-dependent groups account for 99 of 126 points (79\%), while the four reading-only groups contribute only 27. DocVQA shows the same pattern, losing most in table/list questions ($-109$), layout questions ($-84$), and free-text extraction~($-89$).

\subsection{Failure Modes}

We look at the 172 examples where \basevlm{} gets the answer right but \adaptdistill{} does not, and find three patterns.

\paragraph{The model corrects text into real words.}
The language prior turns what the model sees into plausible words, even when the original text is not a real word.
This shows up most in Non-Semantic Text Recognition, where \adaptdistill{} fails on 11 of 50 items (vs.\ 4 for \basevlm{}).
For example, \texttt{PEAEC} becomes \textit{peace}, \texttt{meLtiid} becomes \textit{meltid} (dropping a letter to form a near-word), and \texttt{eeorGg} collapses to \textit{gg}.
In handwriting recognition, the characters are harder to read, which gives the language prior more room to step in.
\texttt{beer} becomes \textit{6er} (a digit replaces an ambiguous stroke), \texttt{both} becomes \textit{6.4}, and \texttt{soul} becomes \textit{saul}.
\basevlm{} reads these correctly because it has a weaker language prior.
\adaptdistill{} fails on 20 of 50 handwriting items, but 11 of those are shared failures where \basevlm{} also gets it wrong, so distillation causes about half the handwriting drop (9 items).

\paragraph{The model reads from the wrong spot.}
In Key Information Extraction, the model often reads from a nearby but wrong part of the document.
When the expected answer is a receipt date (\textit{02/02/2018}), the model returns the store hours (\textit{mon--sun: 1000 hrs -- 2200 hrs}).
When the expected date is \textit{06/03/18}, it returns a timestamp from the next field (\textit{08/03/18 18:04}).
When the expected total is \textit{14.20}, it returns a subtotal from a different row (\textit{13.40}).
The answers are real text from the document, just from the wrong place.
The stronger language prior makes the model favor fluent answers over finding the right field.

\paragraph{The model paraphrases instead of copying.}
The model rewrites what it reads instead of copying it exactly.
Numbers lose precision (\textit{5.04} becomes \textit{5}, \textit{25.9} becomes \textit{25}).
Descriptions get shortened (\textit{21--49 years of age} becomes \textit{21--49 years}).
Values get swapped with nearby ones (\textit{\$506 million} becomes \textit{193,690} from a different table cell).
Names get paraphrased (\textit{program staff} becomes \textit{nutrition health director}).
These are not errors in finding the right spot. The model sees the right region but rephrases what it reads, which is consistent with the teacher being trained on abstractive text.

\newpage
\section{Expanded \& Ablation Results}
\label{app:ablation}

\subsection{Full Results}
\label{tab:full-result}

\begin{table}[H]
\centering
\caption{Full benchmark results across all variants. Columns grouped by training strategy.}
\label{tab:main-expanded}
\setlength{\tabcolsep}{3pt}
\scriptsize
\begin{tabular}{l c cc cc ccc}
\toprule
& & \multicolumn{2}{c}{\cellcolor{lightgray}\textbf{Fine-tuning}} 
& \multicolumn{2}{c}{\cellcolor{lightyellow}\textbf{Uniform KD}} 
& \multicolumn{3}{c}{\cellcolor{lightblue}\textbf{Selective KD}} \\
\cmidrule(lr){3-4}\cmidrule(lr){5-6}\cmidrule(lr){7-9}
\textbf{Task} & \vanillavlm{} &
\cellcolor{lightgray}\texttt{nanoVLM-full} &
\cellcolor{lightgray}\texttt{nanoVLM-lang} &
\cellcolor{lightyellow}\texttt{distill-full} &
\cellcolor{lightyellow}\texttt{distill-lang} &
\cellcolor{lightblue}\texttt{LinguDistill} &
\cellcolor{darkblue}\texttt{highKD} &
\cellcolor{darkblue}\texttt{lowKD} \\
\midrule

\multicolumn{9}{l}{\textit{Language and knowledge-heavy}} \\
AI2D & 0.440 & 0.416 & 0.413 & 0.490 & 0.502 & 0.507 & 0.492 & 0.511 \\
COCO 2017 Captioning2017\_cap & 0.800 & 0.673 & 0.742 & 0.843 & 0.787 & 0.866 & 0.876 & 0.842 \\
ScienceQA & 0.590 & 0.592 & 0.581 & 0.650 & 0.679 & 0.676 & 0.676 & 0.670 \\
ARC Easy & 0.605 & 0.540 & 0.542 & 0.548 & 0.601 & 0.621 & 0.614 & 0.627 \\
ARC Challenge & 0.322 & 0.279 & 0.281 & 0.270 & 0.296 & 0.318 & 0.315 & 0.314 \\
HellaSwag & 0.405 & 0.326 & 0.329 & 0.345 & 0.376 & 0.394 & 0.392 & 0.393 \\

\midrule
\multicolumn{9}{l}{\textit{Document and OCR}} \\
DocVQA\_val & 0.769 & 0.767 & 0.766 & 0.640 & 0.709 & 0.740 & 0.720 & 0.747 \\
InfographicVQA\_val & 0.357 & 0.282 & 0.272 & 0.290 & 0.318 & 0.330 & 0.320 & 0.322 \\
OCRBench & 0.760 & 0.726 & 0.722 & 0.452 & 0.510 & 0.600 & 0.598 & 0.601 \\

\midrule
\multicolumn{9}{l}{\textit{General multimodal}} \\
RealWorldQA & 0.523 & 0.500 & 0.488 & 0.449 & 0.494 & 0.490 & 0.500 & 0.484 \\
mme\_cognition & 302 & 229 & 261 & 241 & 308 & 300 & 247 & 265 \\
mme\_perception & 1259 & 1027 & 1070 & 1046 & 1109 & 1173 & 1082 & 1163 \\
MMMU\_val & 0.320 & 0.310 & 0.302 & 0.283 & 0.288 & 0.283 & 0.293 & 0.288 \\
MMStar & 0.360 & 0.359 & 0.358 & 0.350 & 0.355 & 0.349 & 0.348 & 0.356 \\

\bottomrule
\end{tabular}
\end{table}

\subsection{Data Subset Ablation}
\label{tab:data-ablation}

\begin{table}[H]
\centering
\caption{Effect of data subset selection across all tasks.}
\label{tab:data-ablation}
\small
\begin{tabular}{lccc}
\toprule
Task & \vanillavlm{} & \cellcolor{lightgray}\texttt{nanoVLM-full} & \cellcolor{lightgray}\texttt{nanoVLM-lang} \\
\midrule
\multicolumn{4}{l}{\textit{Language and knowledge-heavy}} \\
AI2D & 0.44 & 0.416 & 0.413 \\
COCO 2017 Captioning & 0.80 & 0.673 & 0.742 \\
ScienceQA & 0.59 & 0.592 & 0.581 \\
ARC Easy & 0.605 & 0.540 & 0.542 \\
HellaSwag & 0.405 & 0.326 & 0.329 \\
\midrule
\multicolumn{4}{l}{\textit{Document and OCR}} \\
DocVQA & 0.769 & 0.767 & 0.766 \\
InfographicVQA & 0.357 & 0.282 & 0.272 \\
OCRBench & 0.760 & 0.726 & 0.722 \\
\midrule
\multicolumn{4}{l}{\textit{General multimodal}} \\
RealWorldQA & 0.523 & 0.500 & 0.488 \\
MME Cognition & 302 & 229 & 261 \\
MME Perception & 1259 & 1027 & 1070 \\
MMMU & 0.32 & 0.31 & 0.302 \\
MMStar & 0.36 & 0.359 & 0.358 \\
\bottomrule
\end{tabular}
\end{table}

\subsection{Uniform vs Selective Distillation Ablation}

\begin{table}[H]
\centering
\caption{Uniform vs selective distillation across all tasks.}
\label{tab:kd-ablation}
\small
\begin{tabular}{lccc}
\toprule
Task & \cellcolor{lightyellow}\texttt{distill-full} & \cellcolor{lightyellow}\texttt{distill-lang} & \cellcolor{lightblue}\texttt{LinguDistill} \\
\midrule
\multicolumn{4}{l}{\textit{Language and knowledge-heavy}} \\
AI2D & 0.490 & 0.502 & 0.507 \\
COCO 2017 Captioning & 0.843 & 0.787 & 0.866 \\
ScienceQA & 0.650 & 0.679 & 0.676 \\
ARC Easy & 0.548 & 0.601 & 0.621 \\
HellaSwag & 0.345 & 0.376 & 0.394 \\
\midrule
\multicolumn{4}{l}{\textit{Document and OCR}} \\
DocVQA & 0.640 & 0.709 & 0.740 \\
InfographicVQA & 0.290 & 0.318 & 0.330 \\
OCRBench & 0.452 & 0.510 & 0.600 \\
\midrule
\multicolumn{4}{l}{\textit{General multimodal}} \\
RealWorldQA & 0.450 & 0.494 & 0.490 \\
MME Cognition & 241 & 308 & 300 \\
MME Perception & 1046 & 1109 & 1173 \\
MMMU & 0.283 & 0.288 & 0.283 \\
MMStar & 0.350 & 0.355 & 0.349 \\
\bottomrule
\end{tabular}
\end{table}

\subsection{Distillation Hyperparameters Ablation}

\begin{table}[H]
\centering
\caption{Effect of distillation strength across all tasks.}
\label{tab:temp-ablation}
\small
\begin{tabular}{lccc}
\toprule
Task & \cellcolor{lightblue}\texttt{LinguDistill} & \cellcolor{darkblue}\texttt{highKD} & \cellcolor{darkblue}\texttt{lowKD} \\
\midrule
\multicolumn{4}{l}{\textit{Language and knowledge-heavy}} \\
AI2D & 0.507 & 0.492 & 0.511 \\
COCO 2017 Captioning & 0.866 & 0.876 & 0.842 \\
ScienceQA & 0.676 & 0.676 & 0.670 \\
ARC Easy & 0.621 & 0.614 & 0.627 \\
HellaSwag & 0.394 & 0.392 & 0.393 \\
\midrule
\multicolumn{4}{l}{\textit{Document and OCR}} \\
DocVQA & 0.740 & 0.720 & 0.747 \\
InfographicVQA & 0.330 & 0.320 & 0.322 \\
OCRBench & 0.600 & 0.598 & 0.601 \\
\midrule
\multicolumn{4}{l}{\textit{General multimodal}} \\
RealWorldQA & 0.490 & 0.500 & 0.484 \\
MME Cognition & 300 & 247 & 265 \\
MME Perception & 1173 & 1082 & 1163 \\
MMMU & 0.283 & 0.293 & 0.288 \\
MMStar & 0.349 & 0.348 & 0.356 \\
\bottomrule
\end{tabular}
\end{table}

\end{document}